\documentclass[11pt]{article}
\usepackage{eacl2017}
\usepackage{times}
\usepackage{url}
\usepackage{latexsym}

\eaclfinalcopy

\usepackage{xspace}
\usepackage{mathtools}
\usepackage{amsmath}
\usepackage{amssymb}
\usepackage[usenames,dvipsnames]{color}
\usepackage{multirow}
\usepackage{tikz}
\usepackage{pifont}
\usepackage{enumitem}% http://ctan.org/pkg/enumitem
\usetikzlibrary{positioning}
\usetikzlibrary{calc}
\usetikzlibrary{fit}
\usetikzlibrary{decorations.text}

\newcommand{\eg}{e.g.\xspace}
\newcommand{\ie}{i.e.\xspace}

\newcommand{\system}[1]{\texttt{#1}\xspace}

\newcommand{\secref}[1]{Section~\ref{#1}\xspace}
\newcommand{\tabref}[2][]{Table#1~\ref{#2}\xspace}
\newcommand{\figref}[2][]{Figure#1~\ref{#2}\xspace}

\newcommand{\bb}[1]{\mathbb{#1}}
\newcommand{\R}{\bb{R}}
\newcommand{\mat}[2][]{\boldsymbol{#2}_{#1}}
\newcommand{\matfwd}[2][]{\overrightarrow{\boldsymbol{#2}}_{#1}}

\renewcommand{\vec}[2][]{\boldsymbol{#2}^{#1}}
\newcommand{\vecfwd}[2][]{\overrightarrow{\boldsymbol{#2}}^{#1}}
\newcommand{\vecrev}[2][]{\overleftarrow{\boldsymbol{#2}}^{#1}}

\newcommand{\T}{\mathstrut\scriptscriptstyle\top}
\newcommand{\tran}{^{\T}}

\newcommand{\myparagraph}[1]{\vspace{0.6\baselineskip}\noindent{\textbf{#1}}~}

\title{A Language-independent and Compositional Model for Personality Trait Recognition from Short Texts}

\author{Fei Liu \\\And
  Julien Perez \\
  Xerox Research Center Europe\\
  {\tt firstname.lastname@xrce.xerox.com} \\\And
  Scott Nowson \\
}

\date{}

\begin{document}
\maketitle
\begin{abstract}
Many methods have been used to recognise author personality traits from text, typically combining linguistic feature engineering with shallow learning models, \eg linear regression or Support Vector Machines. This work uses deep-learning-based models and atomic features of text -- the characters -- to build hierarchical, vectorial word and sentence representations for trait inference. This method, applied to a corpus of tweets, shows state-of-the-art performance across five traits and three languages (English, Spanish and Italian) compared with prior work in author profiling.
The results, supported by preliminary visualisation work, are encouraging for the ability to detect complex human traits.
%from either a collection of sentences or even single ones.
\end{abstract}

\section{Introduction}
\label{(sec:intro)}

Techniques falling under the umbrella of ``deep-learning'' are increasingly commonplace in the space of Natural Language Processing (NLP) \cite{Manning:2016}. Such methods have been applied to a number of tasks from part-of-speech-tagging \cite{Ling+:2015,Huang+:2015} to sentiment analysis \cite{Socher+:2013,Kalchbrenner+:2014,Kim:2014}. Essentially, each of these tasks is concerned with learning representations of language at different levels. The work we outline here is no different in essence, though we choose perhaps the highest level of representation -- that of the author of a given text rather than the text itself.
This task, modelling people from their language, is one built on the long-standing foundation that language use is known to be influenced by sociodemographic characteristics such as gender and personality~\cite{tannen90,pennebaker2003}. The study of personality traits in particular is supported by the notion that they are considered temporally stable \cite{matthews2003}, and thus our modelling ability is enriched by the acquisition of more data over time.

Computational personality recognition, and its broader applications, 
%is gaining increasing traction in academic circles with workshops such as PEOPLES\footnote{Workshop on Computational Modeling of People's Opinions, Personality and Emotions in Social media, \url{http://malvinanissim.github.io/PEOPLES/}} at COLING 2016.
is becoming of increasing interest with workshops exploring the topic \cite{celli2014workshop,empire2014}. The addition of personality traits in the PAN Author Profiling challenge at CLEF in 2015~\cite{rangel:2015} is further evidence. 
%\felix{Add the COLING personality workshop here perhaps?} There is also interest on the commercial side, companies such as Receptiviti and Juji~\footnote{http://www.receptiviti.ai/ and https://juji.io/ respectively} are building strong business cases that psychometric analysis is a core value-add to any customer-driven application.
Much prior literature in this field has used some variation of enriched bag-of-words; \eg the “Open vocabulary” approach ~\cite{schwartz2013}. This is understandable as exploring the relationship between word use and traits has delivered significant insight into aspects of human behaviour~\cite{pennebaker2003}. Different levels of representation of language have been used such as syntactic, semantic, and higher-order such as the psychologically-derived lexica of the Linguistic Inquiry and Word Count (LIWC) tool~\cite{liwc2015}.

One drawback of this bag-of-linguistic-features approach is that considerable effort can be spent on feature engineering. Moreover, such linguistic features are mostly language-dependent, such as LIWC \cite{liwc2015}, making the adoptation to multi-lingual models more time-comsuming. Another issue is an unspoken assumption that these features, like the traits to which they relate, are similarly stable: the same language features always indicate the same traits. However, this is not the case. As Nowson and Gill~\shortcite{nowson2014} have shown, the relationship between language and personality is not consistent across all forms of communication
%\footnote{The use of negative emotion language (in particular, words relating to 'anger' is a strong indicator of Extraversion in conversational data, but not in the medium of video blogging.}
 – 
 %there are many other factors of context to consider, 
 the relationship is more complex. 

In order to better explore this complexity in this work we propose a novel deep-learning feature-engineering-free modelisation of the problem of personality trait recognition, making the model language independent and enablling it to work in various languages without the need to create language-specific linguistic features. The task is framed as one of supervised sequence regression based on a joint atomic representation of the text: specifically on the character and word level. In this context, we are exploring short texts. Typically, classification of such texts tends to be particularly challenging for state-of-the-art BoW based approaches due, in part, to the noisy nature of such data \cite{Han+:2011}.
%to the specific uses of language of the user driven by the message length limitation of the platforms, \eg 140 characters in Twitter, causing the user-generated text to be noisy and packed with out-of-vocabulary (OOV) words. 
To cope with this 
%the noisy nature of texts from such platforms, 
we propose a novel recurrent and compositional neural network architecture, capable of constructing representations at character, word and sentence level. We believe a latent representation inference based on a parse-free input representation of the text seen as a sequence of characters can balance the bias and variance of such sparse dataset.

The paper is structured as follows: after we consider previous approaches to the task of computational personality recognition, including those which have a deep-learning component, we describe our model. We report on two sets of experiments, the first of which demonstrates the effectiveness of the model in inferring personality for users, while the second reports on the short text level analysis. In both settings, the proposed model achieves state-of-the-art performance across five personality traits and three languages.

%Note that there are many approaches to modelling personality traits and this invention is agnostic of any specific one.  However, for ease of reference and evaluation we use a single model -- the Big 5 -- with which we will ground this proposal.

\section{Related Work}
\label{sec:bg}

%\begin{itemize}
%	\item can also talk about representation level, e.g. \cite{verhoeven2014} for using character level analysis for personality	
%	\ITEM MAYBE WITH MORE SPACE
%\end{itemize}

Early work on computational personality recognition~\cite{argamon2005,nowson2006} used SVM-based approaches and manipulated lexical and grammatical feature sets. 
%The first work on computational personality recognition~\cite{argamon2005} used an SVM-based approach and sets of lexically and grammatically inspired features. In the decade that has followed, there has been little evolution in techniques. 
Today, according to the organisers~\cite{rangel:2015} ``most'' participants to the PAN 2015 Author Profiling task still use a combination of SVM and feature engineering. 
%Ensemble methods have been proposed~\cite{verhoeven2013}, but the base model was still SVM -- the ensemble came from the combination of data from different sources as opposed to models. 
Data labelled with personality data is sparse~\cite{nowson2014} and there has been more interest in reporting novel feature sets. In the PAN task alone\footnote{Due to space consideration we are unable to cite the individual works.} there were features used from multiple levels of representation on language. Surface forms were present in word, lemma and character n-grams, while syntactic features included POS tags and dependency relations. There were some efforts of feature curation, such as analysis of punctuation and emoticon use, along with the use of latent semantic analysis for topic modelling. Another popular feature set is the use of external resources such as LIWC~\cite{liwc2015} which, in this context, represents over 20 years of psychology-based feature engineering. When applied to tweets, however, LIWC requires further cleaning of the data \cite{Receptiviti}.

Deep-learning based approaches to personality trait recognition are, unsurprisingly given the typical size of data sets, relatively few. The model detailed in \newcite{kalghatgi2015} presents a neural network based approach to personality prediction of users. In this model, a Multilayer Perceptron (MLP) takes as input a collection of hand-crafted grammatical and social behavioral features from each user and assigns a label to each of the 5 personality traits. Unfortunately no evaluation of this work, nor details of the dataset, were provided. The work of \newcite{su2016} describes a Recurrent Neural Network (RNN) based system, exploiting the turn-taking of conversation for personality trait prediction. In their work, RNNs are employed to model the temporal evolution of dialog, taking as input LIWC-based and grammatical features. The output of the RNNs is then used for the prediction of personality trait scores of the participants of the conversations. It is worth noting that both works utilise hand-crafted features which rely heavily on domain expertise. Also the focus is on the prediction of trait scores on the user level given all the available text from a user. 
%In contrast, the goal of the approach presented in this paper is the prediction of trait scores from a single short text, arguably a more challenging task considering the limited amount of information. % with prediction on the user level.
In contrast, not only can the approach presented in this paper infer the personality of a user given a collection of short texts, it is also flexible to predict trait scores from a single short text, arguably a more challenging task considering the limited amount of information.

The model we present in \secref{sec:cwm} is inspired by \newcite{Ling+:2015}, who proposed a character-level word representation learning model under the assumption that character sequences are syntactically and semantically informative of the words they compose. Based on a widely used RNN named long short-term memory network (LSTM)~\cite{Hochreiter+:1997}, 
%capable of capturing complex non-local dependencies in sequences, 
the model learns the embeddings of characters and how they can be used to construct words. Topped by a softmax layer at each word, the model was applied to the tasks of language modelling and part-of-speech tagging and successful in improving upon traditional baselines particularly in morphologically rich languages. 
%Not only does the model achieve better performance on both tasks, it also has significantly fewer parameters to learn compared to a word look-up table based approach as the number of different characters is much fewer than the number of different words in a vocabulary. Moreover, the model is able to generate a sensible representation for unseen words. 
Inspired by this, \newcite{Yang+:2016} introduced Hierarchical Attention Networks where the representation of a document is hierarchically built up. They construct the representation of a sentence by processing a sequence of its constituent words using a bi-directional gated recurrent unit (GRU) \cite{Cho+:2014b}. The representations of sentences are in turn processed by another bi-directional GRU at the sentence level to form the representation of the document. The work of \cite{Ling+:2015} provides a way to construct words from their constituent characters (Character to Word, \system{C2W}) while \newcite{Yang+:2016} describe a hierarchical approach to building representations of documents from words to sentences, and eventually to documents (Word to Sentence to Document, \system{W2S2D}). In this work, inspired by the above works, we present a hierarchical model situated between the above two models, connecting characters, words and sentences, and ultimately personality traits (Character to Word to Sentence for Personality Trait, \system{C2W2S4PT}).

\section{Proposed Model}

To motivate our methodology, we review a commonly-used approach to representing sentences and discuss some of its limitations and motivation. Then, we propose the use of a compositional model to tackle the identified problems. 

\subsection{Current Issues and Motivation}
\label{sec:problems}

One classical approach for applying deep learning models to NLP problems involves word lookup tables where words are typically represented by dense real-valued vectors in a low-dimensional space \cite{Socher+:2013,Kalchbrenner+:2014,Kim:2014}. In order to obtain a sensible set of embeddings, a common practice is to train on a large corpus in an unsupervised fashion, \eg Word2Vec \cite{Mikolov+:2013a,Mikolov+:2013b} and GloVe \cite{Pennington+:2014}.
Despite the success in capturing syntactic and semantic information with such word vectors, there are two practical problems with such an approach \cite{Ling+:2015}. First, due to the flexibility of language, previously unseen words are bound to occur regardless of how large the unsupervised training corpus is. The problem is particularly serious for text extracted from social media platforms such as Twitter and Facebook due to the noisy nature of user-generated text -- \eg typos, ad hoc acronyms and abbreviations, phonetic substitutions, and even meaningless strings \cite{Han+:2011}. A naive solution is to map all unseen words to a vector UNK representing the unknown word. Not only does this approach give up critical information regarding the meaning of the unknown words, it is also difficult for the model to generalise to made up words, such as \textit{beautification}, despite the components \textit{beautiful} and \textit{-ification} having been observed. Second, the number of parameters for a model to learn is overwhelmingly large. Assume each word is represented by a vector of $d$ dimensions, the total size of the word lookup table is $d \times |V|$ where $|V|$ is the size of the vocabulary which tends to scale to the order of hundreds and thousands. Again, this problem is even more pronounced in noisier domain such as short text generated by online users. To address the above issues, we adopt a compositional character to word model described in the next section.

From the personality perspective, character-based features have been widely adopted in trait inference, such as character n-grams\cite{Gonzalezgallardo+:2015,Sulea+:2015}, emoticons \cite{nowson2015,Palominogaribay+:2015}, and character flooding \cite{nowson2015,Gimenez+:2015}. Motivated by this and the issues identified above, we propose in the next section a language-independent compositional model that operates hierarchically at the character, word and sentence level, capable of harnessing personality-sensitive signals buried as deep as the character level.

%\subsection{A Recurrent and Compositional Model for Personality Trait Inference and Representation}
%\label{sec:model}
%
%We propose two approaches to address the concerns above, which we will apply to the task of inferring personality traits. First, we propose to sequentially and hierarchically learn the latent representation of characters and words in order to leverage the information embedded in these two fundamental representations of text. Second, we leverage correlations between traits in order to perform multitask learning.

%\subsection{Character Level Motivation}

\subsection{Character to Word to Sentence for Personality Traits}
\label{sec:cwm}

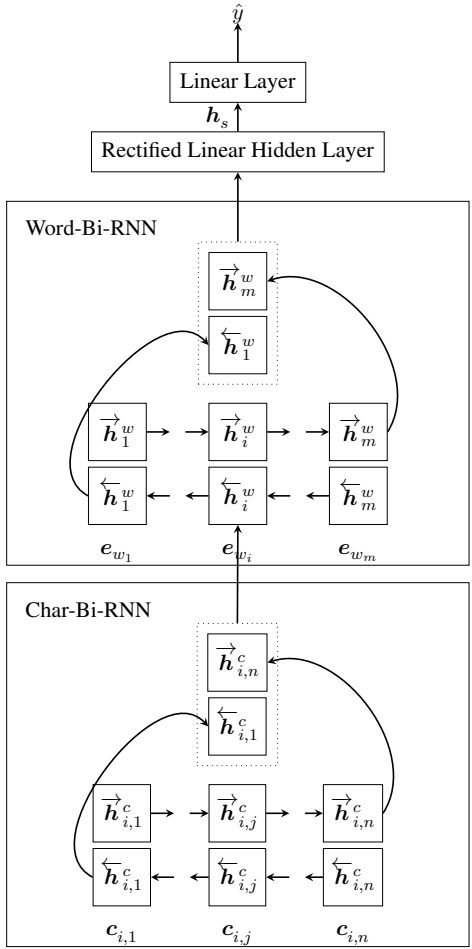
\begin{figure}[tb]
%\vspace{-0.5cm}
\begin{center}
\resizebox{.82\columnwidth}{!}{
\begin{tikzpicture}

%% Char RNN
\node[align=center] at (0.0,0.0) (c_w_i_1) {$\vec{c}_{i,1}$};
%\node[align=center] at (1.5,0.0) (c_w_i_2) {$c_{i,2}$};
\node[align=center] at (2.0,0.0) (c_w_i_3) {$\vec{c}_{i,j}$};
\node[align=center] at (4.0,0.0) (c_w_i_n) {$\vec{c}_{i,n}$};
\node[draw,align=center,minimum height=1.0cm] (hr_i_1) [above=0.2of c_w_i_1] {$\vecrev[c]{h}_{i, 1}$};
\node[draw,align=center,minimum height=1.0cm] (hf_i_1) [above=0.1of hr_i_1] {$\vecfwd[c]{h}_{i, 1}$};

\node[align=center,minimum height=1.0cm] (hr_i_2) [right=0.4of hr_i_1] {};
\node[align=center,minimum height=1.0cm] (hf_i_2) [above=0.1of hr_i_2] {};

\node[draw,align=center,minimum height=1.0cm] (hr_i_3) [above=0.2of c_w_i_3] {$\vecrev[c]{h}_{i, j}$};
\node[draw,align=center,minimum height=1.0cm] (hf_i_3) [above=0.1of hr_i_3] {$\vecfwd[c]{h}_{i, j}$};

\node[align=center,minimum height=1.0cm] (hr_i_4) [right=0.4of hr_i_3] {};
\node[align=center,minimum height=1.0cm] (hf_i_4) [right=0.4of hf_i_3] {};

\node[draw,align=center,minimum height=1.0cm] (hr_i_n) [above=0.2of c_w_i_n] {$\vecrev[c]{h}_{i, n}$};
\node[draw,align=center,minimum height=1.0cm] (hf_i_n) [above=0.1of hr_i_n] {$\vecfwd[c]{h}_{i, n}$};

\node[draw,align=center,minimum height=1.0cm] (hr_i_c) [above=0.5of hf_i_3] {$\vecrev[c]{h}_{i, 1}$};
\node[draw,align=center,minimum height=1.0cm] (hf_i_c) [above=0.1of hr_i_c] {$\vecfwd[c]{h}_{i, n}$};

%\node[draw,dashed,fit=(hr_i_1) (hf_i_1),inner sep=5] {};
%%\node[draw,dashed,fit=(hr_i_2) (hf_i_2),inner sep=5] {};
%\node[draw,dashed,fit=(hr_i_3) (hf_i_3),inner sep=5] {};
%\node[draw,dashed,fit=(hr_i_n) (hf_i_n),inner sep=5] {};

\node[draw,dotted,fit=(hr_i_c) (hf_i_c),inner sep=5] (e_w_i) {};

\node[draw,fit=(hr_i_n) (e_w_i) (hr_i_1),inner sep=20,minimum width=8.0cm] (char_rnn) {};
\node[anchor=north west,shift={(2mm,-2mm)}] (char_rnn_label) at (char_rnn.north west){Char-Bi-RNN};

\path[draw,thick,->,>=stealth] (hf_i_1) to (hf_i_2);
\path[draw,thick,->,>=stealth] (hf_i_2) to (hf_i_3);
\path[draw,thick,->,>=stealth] (hf_i_3) to (hf_i_4);
\path[draw,thick,->,>=stealth] (hf_i_4) to (hf_i_n);

\path[draw,thick,->,>=stealth] (hr_i_n) to (hr_i_4);
\path[draw,thick,->,>=stealth] (hr_i_4) to (hr_i_3);
\path[draw,thick,->,>=stealth] (hr_i_3) to (hr_i_2);
\path[draw,thick,->,>=stealth] (hr_i_2) to (hr_i_1);

\path[draw,thick,->,>=stealth,bend left=100] (hr_i_1.west) to (hr_i_c.west);
\path[draw,thick,->,>=stealth,bend right=70] (hf_i_n.east) to (hf_i_c.east);

%% Word RNN
\node[draw,align=center,minimum height=1.0cm,minimum width=1cm] (hr_i) [above=1.0of char_rnn] {$\vecrev[w]{h}_{i}$};
\node[draw,align=center,minimum height=1.0cm,minimum width=1cm] (hf_i) [above=0.1 of hr_i] {$\vecfwd[w]{h}_{i}$};

\node[align=center,minimum height=1.0cm] (hr_2) [left=0.4of hr_i] {};
\node[align=center,minimum height=1.0cm] (hf_2) [above=0.1of hr_2] {};

\node[draw,align=center,minimum height=1.0cm,minimum width=1cm] (hr_1) [left=0.4of hr_2] {$\vecrev[w]{h}_{1}$};
\node[draw,align=center,minimum height=1.0cm,minimum width=1cm] (hf_1) [above=0.1of hr_1] {$\vecfwd[w]{h}_{1}$};

\node[align=center,minimum height=1.0cm] (hr_4) [right=0.4of hr_i] {};
\node[align=center,minimum height=1.0cm] (hf_4) [above=0.1of hr_4] {};

\node[draw,align=center,minimum height=1.0cm,minimum width=1cm] (hr_m) [right=0.4of hr_4] {$\vecrev[w]{h}_{m}$};
\node[draw,align=center,minimum height=1.0cm,minimum width=1cm] (hf_m) [above=0.1of hr_m] {$\vecfwd[w]{h}_{m}$};

\node[draw,align=center,minimum height=1.0cm,minimum width=1.0cm] (hr_w) [above=0.5of hf_i] {$\vecrev[w]{h}_{1}$};
\node[draw,align=center,minimum height=1.0cm,minimum width=1.0cm] (hf_w) [above=0.1of hr_w] {$\vecfwd[w]{h}_{m}$};

\node[dashed,fit=(hr_1) (hf_1),inner sep=5] (h_1) {};
\node[dashed,fit=(hr_i) (hf_i),inner sep=5] (h_i) {};
\node[dashed,fit=(hr_m) (hf_m),inner sep=5] (h_m) {};
\node[draw,dotted,fit=(hr_w) (hf_w),inner sep=5] (e_s) {};

\node[draw,fit=(hr_m) (e_s) (hr_1),inner sep=20,minimum width=8.0cm] (word_rnn) {};
\node[anchor=north west,shift={(2mm,-2mm)}] (word_rnn_label) at (word_rnn.north west){Word-Bi-RNN};

%\path[draw,thick,->,>=stealth,bend left=100] (e_w_i.north) to (h_i.south);
\path[draw,thick,->,>=stealth] (hf_1) to (hf_2);
\path[draw,thick,->,>=stealth] (hf_2) to (hf_i);
\path[draw,thick,->,>=stealth] (hf_i) to (hf_4);
\path[draw,thick,->,>=stealth] (hf_4) to (hf_m);

\path[draw,thick,->,>=stealth] (hr_m) to (hr_4);
\path[draw,thick,->,>=stealth] (hr_4) to (hr_i);
\path[draw,thick,->,>=stealth] (hr_i) to (hr_2);
\path[draw,thick,->,>=stealth] (hr_2) to (hr_1);

\path[draw,thick,->,>=stealth,bend left=100] (hr_1.west) to (hr_w.west);
\path[draw,thick,->,>=stealth,bend right=70] (hf_m.east) to (hf_w.east);
\draw [thick,->,>=stealth] plot coordinates { (e_w_i.north) (hr_i.south)};

\node[align=center] (w_1) [below=0.2of hr_1] {$\vec{e}_{w_1}$};
\node[align=center] (w_i) [below=0.2of hr_i] {$\vec{e}_{w_i}$};
\node[align=center] (w_m) [below=0.2of hr_m] {$\vec{e}_{w_m}$};

%% MLP Regression
\node[align=center,draw,inner sep=5] [above=0.5 of word_rnn] (nonlinear) {Rectified Linear Hidden Layer};
\node[align=center,draw,inner sep=5] [above=0.5 of nonlinear] (linear) {Linear Layer};
\draw [thick,->,>=stealth] plot coordinates { (e_s.north) (nonlinear.south)};
\node[align=center,inner sep=5] [above=0.5 of linear] (y_pred) {$\hat{y}$};
%\draw [thick,->,>=stealth,postaction={decorate,decoration={text along path,text align=center,text={|\sffamily|A}}}] plot [smooth, tension=1] coordinates { (nonlinear.north)  (linear.south)};
\path[thick,->,>=stealth] (nonlinear.north) edge node[anchor=center, left, midway] {$\vec{h}_s$} (linear.south);
\draw [thick,->,>=stealth] plot [smooth, tension=1] coordinates { (linear.north)  (y_pred.south)};
\end{tikzpicture}
}
\end{center}
%\vspace{-1.0cm}
\caption{\label{fig:c2w}Illustration of the \system{C2W2S4PT} model. Dotted boxes indicate concatenation.}
\end{figure}

To address the problems identified in \secref{sec:problems}, we propose to extend the compositional character to word model first introduced by \newcite{Ling+:2015} wherein the representation of each word is constructed, via a character-level bi-directional RNN (Char-Bi-RNN), from its constituent characters. 
%An illustration of this model is shown in \figref{fig:c2w}. 
The constructed word vectors are then fed to another layer of word-level Bi-RNN (Word-Bi-RNN) and a sentence is represented by the concatenation of the last and first hidden states of the forward and backward Word-RNNs respectively. Eventually, a feedforward neural network takes as input the representation of a sentence and returns a scalar as the prediction for a specific personality trait. Thus, we name the model \system{C2W2S4PT} (Character to Word to Sentence for Personality Traits) which is illustrated in \figref{fig:c2w}. Specifically, suppose we have a sentence $s$ consisting of a sequence of words $\{w_1, w_2, \dots, w_i, \dots, w_m\}$. We define a function $\textrm{c}(w_i, j)$ which takes as input a word $w_i$, together with an index $j$ and returns the one-hot vector representation of the $j^{\textrm{th}}$ character of the word $w_i$. Then, to get the embedding $\vec{c}_{i, j}$ of the character, 
%we transform the one-hot vector to a $d$-dimensional vector using a matrix $\mat[c]{E} \in \R^{d \times |C|}$ where $|C|$ is the size of the character vocabulary. 
we transform $\textrm{c}(w_i, j)$ by: $\vec{c}_{i, j} = \mat[c]{E}\textrm{c}(w_i, j)$ where $\mat[c]{E} \in \R^{d \times |C|}$ and $|C|$ is the size of the character vocabulary. 
Next, in order to construct the representation of word $w_i$, the sequence of character embeddings $\{\vec{c}_{i, 1}, \dots, \vec{c}_{i, n}\}$ is taken as input to the Char-Bi-RNN (assuming $w_i$ is comprised of $n$ characters). In this work, we employ GRU 
% already introduced GRU in Related Work
%recently introduced by \newcite{Cho+:2014b}, 
as the recurrent unit in the Bi-RNNs, given that recent studies indicate that GRU achieves comparable, if not better, results to LSTM \cite{Chung+:2014,Kumar+:2015,Jozefowicz+:2015}.\footnote{We performed additional experiments which confirmed this finding. Therefore due to space considerations, we do not report results using LSTMs here.} 
Concretely, the forward pass of the Char-Bi-RNN is carried out using the following:
%\vspace*{-3mm}
\begin{align}
\vecfwd[c]{z}_{i, j} &= \sigma(\matfwd[z]{W}^{c}\vec{c}_{i, j} + \matfwd[hz]{U}^{c}\vecfwd[c]{h}_{i, j-1} + \vecfwd[c]{b}_{z}) \\
\vecfwd[c]{r}_{i, j} &= \sigma(\matfwd[r]{W}^{c}\vec{c}_{i, j} + \matfwd[hr]{U}^{c}\vecfwd[c]{h}_{i, j-1} + \vecfwd[c]{b}_{r}) \\
\vecfwd[c]{\tilde{h}}_{i, j} &= \tanh(\matfwd[h]{W}^{c}\vec{c}_{i, j} + \vecfwd[c]{r}_{i, j}\odot\matfwd[hh]{U}^{c}\vecfwd[c]{h}_{i, j-1} + \vecfwd[c]{b}_{h}) \\
\vecfwd[c]{h}_{i, j} &= \vecfwd[c]{z}_{i, j}\odot\vecfwd[c]{h}_{i, j-1} + (1 - \vecfwd[c]{z}_{i, j})\odot\vecfwd[c]{\tilde{h}}_{i, j}
\end{align}
%\begin{equation*}
%\begin{aligned}[c]
%\vecfwd[c]{h}_{i, j} &= \grufwd(\vec{c}_{i, j}, \vecfwd[c]{h}_{i, j-1})
%\end{aligned}
%\qquad\qquad
%\begin{aligned}[c]
%\vecrev[c]{h}_{i, j} &= \grurev(\vec{c}_{i, j}, \vecrev[c]{h}_{i, j+1})
%\end{aligned}
%\end{equation*}
where $\odot$ is the element-wise product, $\matfwd[z]{W}^{c}, \matfwd[r]{W}^{c}, \matfwd[h]{W}^{c}, \matfwd[hz]{U}^{c}, \matfwd[hr]{U}^{c}, \matfwd[hh]{U}^{c}$ are the parameters for the model to learn, and $\vecfwd[c]{b}_{z}, \vecfwd[c]{b}_{r}, \vecfwd[c]{b}_{h}$ the bias terms. The backward pass, the hidden state of which is symbolised by $\vecrev[c]{h}_{i, j}$, is performed similarly, although with a different set of GRU weight matrices and bias terms. 
It should be noted that both the forward and backward Char-RNN share the same character embeddings. Ultimately, $w_i$ is represented by the concatenation of the last and first hidden states of the forward and backward Char-RNNs: $\vec{e}_{w_i} = [\vecfwd[c]{h}_{i, n};\vecrev[c]{h}_{i, 1}]\tran$.
%\begin{equation*}
%\vec{e}_{w_i} = \begin{bmatrix}\vecfwd[c]{h}_{i, n}\\\vecrev[c]{h}_{i, 1}\end{bmatrix} 
%\end{equation*}
Once all the word representations $\vec{e}_{w_i}$ for $i \in [1, n]$ have been constructed from their constituent characters, they are then processed by the Word-Bi-RNN, similar to Char-Bi-RNN but on word level with word rather than character embeddings:
\begin{align}
\vecfwd[w]{z}_i &= \sigma(\matfwd[z]{W}^{w}\vec{e}_{w_i} + \matfwd[hz]{U}^{w}\vecfwd[w]{h}_{i-1} + \vecfwd[w]{b}_{z}) \\
\vecfwd[w]{r}_i &= \sigma(\matfwd[r]{W}^{w}\vec{e}_{w_i} + \matfwd[hr]{U}^{w}\vecfwd[w]{h}_{i-1} + \vecfwd[w]{b}_{r}) \\
\vecfwd[w]{\tilde{h}}_i &= \tanh(\matfwd[h]{W}^{w}\vec{e}_{w_i} + \vecfwd[w]{r}_i\odot\matfwd[hh]{U}^{w}\vecfwd[w]{h}_{i-1} + \vecfwd[w]{b}_{h}) \\
\vecfwd[w]{h}_i &= \vecfwd[w]{z}_i\odot\vecfwd[w]{h}_{i-1} + (1 - \vecfwd[w]{z}_i)\odot\vecfwd[w]{\tilde{h}}_i
\end{align}
%\begin{equation*}
%\begin{aligned}[c]
%\vecfwd[w]{h}_i &= \grufwd(\vec{e}_{w_i}, \vecfwd[w]{h}_{i-1})
%\end{aligned}
%\qquad\qquad
%\begin{aligned}[c]
%\vecrev[w]{h}_i &= \grurev(\vec{e}_{w_i}, \vecfwd[w]{h}_{i+1})
%\end{aligned}
%\end{equation*}
where $\matfwd[z]{W}^{w}, \matfwd[r]{W}^{w}, \matfwd[h]{W}^{w}, \matfwd[hz]{U}^{w}, \matfwd[hr]{U}^{w}, \matfwd[hh]{U}^{w}$ are the parameters for the model to learn, and $\vecfwd[w]{b}_{z}, \vecfwd[w]{b}_{wr}, \vecfwd[w]{b}_{h}$ the bias terms. 
In a similar fashion to how a word is represented, we construct the sentence embedding by concatenation: $\vec{e}_s = [\vecfwd[w]{h}_{m};\vecrev[w]{h}_{1}]\tran$.
%\begin{equation*}
%\vec{e}_{s} = \begin{bmatrix}\vecfwd[w]{h}_{m}\\\vecrev[w]{h}_{1}\end{bmatrix}
%\end{equation*}
Lastly, to estimate the score for a particular personality trait, we top the Word-Bi-RNN with an MLP which takes as input the sentence embedding $\vec{e}_s$ and returns the estimated score $\hat{y}_{s}$: 
%$\vec{h}_{s} = \textrm{ReLU}(\mat[eh]{W}\vec{e}_{s} + \vec{b}_{h})$ and then $\hat{y}_{s} = \mat[hy]{W}\vec{h}_{s} + b_{y}$
\begin{align}
\vec{h}_{s} &= \max(0, \mat[eh]{W}\vec{e}_{s} + \vec{b}_{h}) \\
\hat{y}_{s} &= \mat[hy]{W}\vec{h}_{s} + b_{y}
\end{align}
%\begin{equation*}
%\begin{aligned}[c]
%\vec{h}_{s} &= \textrm{ReLU}(\mat[eh]{W}\vec{e}_{s} + \vec{b}_{h})
%\end{aligned}
%\qquad\qquad
%\begin{aligned}[c]
%\hat{y}_{s} &= \mat[hy]{W}\vec{h}_{s} + b_{y}
%\end{aligned}
%\end{equation*}
where $\textrm{ReLU}$ is the REctified Linear Unit defined as $\textrm{ReLU}(x) = \max(0, x)$, $\mat[eh]{W}, \mat[hy]{W}$ the parameters for the model to learn, $\vec{b}_{h}, b_{y}$ the bias terms, and $\vec{h}_{s}$ the hidden representation of the MLP. All the components in the model are jointly trained with \textit{mean square error} being the objective function: 
%$L(\theta) = \frac{1}{n}\sum_{i = 1}^{n}(y_{s_i} - \hat{y}_{s_i})^2$
\begin{equation}
L(\theta) = \frac{1}{n}\sum_{i = 1}^{n}(y_{s_i} - \hat{y}_{s_i})^2
\end{equation}
where $y_{s_i}$ is the ground truth personality score of sentence $s_i$ and $\theta$ the collection of all embedding and weight matrices and bias terms for the model to learn. 
Note that no language-dependent component is present in the proposed model.

\section{Experiments and Results}

We report two sets of experiments: 
%the first designed to evaluate the performance of the model on individual short texts; the second an adaptation of the evaluation metric applied at the user level by aggregating the scores assigned to each short text, intended to enable comparison with the current state of the art.
%the first designed to evaluate the performance of the proposed model against other feature-engineering-free approaches on individual short texts; the second an comparison at the user level between our model and current state-of-the-art models which rely on linguistic features. We show that in both settings (with or without feature engineering), our proposed model achieves better results across two languages.
the first a comparison at the user level between our feature-engineering-free  and language-independent approach and current state-of-the-art models which rely on linguistic features; the second designed to evaluate the performance of the proposed model against other feature-engineering-free approaches on individual short texts. We show that in both settings, \ie, against models with or without feature engineering, our proposed model achieves better results across two languages (English and Spanish) and is equally competitive in Italian.

%In this section, we first describe the dataset. Then the experimental setup is detailed, followed by the results with comparison to previous state-of-the-art methods. Lastly, we present the 

\subsection{Dataset and Data Preprocessing}
\label{sec:data}
We use the English, Spanish and Italian data from the PAN 2015 Author Profiling task dataset~\cite{rangel:2015}, collected from Twitter and consisting of $14,166$ English (EN), $9,879$ Spanish (ES) and $3,687$ Italian (IT) tweets (from $152$, $110$ and $38$ users respectively). Due to space constraints and the limited size of the data, the Dutch dataset is not included. For each user there is a set of tweets (average $n=100$) and gold standard personality labels. The five trait labels, scores between -0.5 and 0.5, are calculated following the author's self-assessment responses to the short Big 5 test, BFI-10~\cite{rammstedt2007} which is the most widely accepted and exploited scheme for personality recognition and has the most solid grounding in language \cite{Poria+:2013}.
%Each score is between -0.5 and 0.5, and we consider each short text to inherit to trait scores of the author as text-level annotations.

%\subsection{Data Preprocessing}
In our experiments, each tweet is tokenised using Twokenizer \cite{Owoputi+:2013}, in order to preserve hashtag-preceded topics and user mentions. Unlike the majority of the language used in a tweet, URLs and mentions are used for their targets, and not their surface forms. Therefore each text is normalised by mapping these features to single characters (\eg, \textit{@username} $\rightarrow$ \textit{@}, \textit{http://t.co/} $\rightarrow$ \textit{\string^}). Thus we limit the risk of modelling, say, character usage which was not directly influenced by the personality of the author.

%We believe that it is the occurrences of user mentions and URLs that are informative rather than the actual text of usernames and URLs \felix{need references???} \scott{we can probably cite some work which says references to other people is an indicator. or maybe we don't mention it, because that's borderline assuming features}.
% Cite from the overview paper, Section 4, Preprocessing

\subsection{Evaluation Method}
%The submissions to the PAN 2015 task were evaluated by the organisers on unseen data. As this is not available, 
Due to the unavailability of the test corpus -- withheld by the PAN 2015 organisers -- we compare the $k$-fold cross-validation performance ($k = 5$ or $10$) on the available dataset. Performance is measured using Root Mean Square Error (RMSE) on either the tweet level or user level depending on the granularity of the task: 
$RMSE_{tweet} = \sqrt{\frac{\sum_{i=1}^{T}(y_{s_{i}} - \hat{y}_{s_{i}})^2}{n}}$ and $RMSE_{user} = \sqrt{\frac{\sum_{i=1}^{U}(y_{user_i} - \hat{y}_{user_{i}})^2}{n}}$
%\begin{align}
%RMSE_{tweet} &= \sqrt{\dfrac{\sum_{i=1}^{T}(y_{s_{i}} - \hat{y}_{s_{i}})^2}{n}}\\
%RMSE_{user} &= \sqrt{\dfrac{\sum_{i=1}^{U}(y_{user_i} - \hat{y}_{user_{i}})^2}{n}}
%\end{align}
%\begin{equation*}
%\begin{aligned}[c]
%RMSE_{tweet} &= \sqrt{\dfrac{\sum_{i=1}^{T}(y_{s_{i}} - \hat{y}_{s_{i}})^2}{n}}
%\end{aligned}
%\qquad
%\begin{aligned}[c]
%RMSE_{user} &= \sqrt{\dfrac{\sum_{i=1}^{U}(y_{user_i} - \hat{y}_{user_{i}})^2}{n}}
%\end{aligned}
%\end{equation*}
where $T$ and $U$ are the total numbers of tweets and users in the corpus, $y_{s_i}$ and $\hat{y}_{s_{i}}$ the true and estimated personality trait score of the $i^{\textrm{th}}$ tweet, similarly $y_{user_i}$ and $\hat{y}_{user_{i}}$ are their user-level counterparts. 
Each tweet in the dataset inherits the same five trait scores as assigned to the author from whom they were drawn.
%Note that in the dataset utilized in this work, each user is assigned a single score for a particular personality trait and every tweet collected from the same account receives the same five personality trait assignments as its author. 
$\hat{y}_{user_{i}} = \frac{1}{T_{i}}\sum_{j=1}^{T_{i}}\hat{y}_{s_j}$ where $T_{i}$ refers to the total number of tweets of $user_i$. In \secref{sec:userlevel} and \secref{sec:tweetlevel}, we present the results measured at the user and tweet level using $RMSE_{user}$ and $RMSE_{tweet}$ respectively. It is important to note that, to enable direct comparison, we use exactly the same dataset and evaluation metric $RMSE_{user}$ as in the works of \cite{Sulea+:2015,Mirkin+:2015,nowson2015}.

\subsection{Personality Trait Prediction at User Level}
\label{sec:userlevel}

We test the proposed models on the dataset described in \secref{sec:data} and train our model to predict the personality trait scores based purely on the text without additional features supplied.
To demonstrate the effectiveness of the proposed model, we evaluate the performance on the user level against models incorporating linguistic and psychologically motivated features. This allows us to directly compare the performance of current state-of-the-art models and \system{C2W2S4PT}.
%Datasets with personality labels are typically used for the task of user profiling, not trait prediction from single texts. Thus the performance reported in \tabref{tbl:performancetweetlevel} is not directly comparable with prior results. In prior uses of this dataset in particular, the goal is to predict personality trait scores on user level given a number of texts. In order to enable such a comparison, we re-run our experiments with the same hyper-parameter configuration as described in \secref{sec:tweetlevel} reporting the user scores as the arithmetic mean of the estimated scores assigned to each of their tweets. 
%to compare the performance between our model and the previous state-of-the-art models, we conduct experiments in this setting and compute the personality trait scores for a user by taking the mean of the estimated scores assigned to each tweet of the user by our model. 
%In \tabref{tbl:performancetweetlevel}, the joint models demonstrate equally competitive performance as the models trained on each personality trait individually while taking similar time to train. We therefore only present the performance of the joint model on user level.
%Due to the unavailability of the test corpus, we compare the $k$-fold cross-validation performance ($k = 5$ or $10$) on the training dataset. 
For $5$-fold cross-validation, we compare to the tied-highest ranked (under evaluation conditions in EN, ranked $7^{\textrm{th}}$ and $4^{\textrm{th}}$ in ES and IT) of the PAN 2015 submissions~\cite{Sulea+:2015}.\footnote{Cross-validation $RMSE_{user}$ performance is not reported for the other top system \cite{alvarezcarmona:2015}.} For $10$-fold cross-validation, we similarly choose the work by ranking and metric reporting~\cite{nowson2005} (ranked $9^{\textrm{th}}$, $6^{\textrm{th}}$ and $8^{\textrm{th}}$ in EN, ES and IT). 
As here, these works predicted scores on text level, and averaged for each user. Therefore, we include subsequent work which reports results on concatenated tweets -- a single document per user~\cite{Mirkin+:2015}. 
For each language, we also show the most straightforward baseline \system{Average Baseline} which assigns the average of all the scores to each user. 
\system{C2W2S4PT} is trained with Adam \cite{Kingma+:2014} and hyper-parameters: 
$\mat[c]{E} \in \R^{50 \times |C|}$, 
$\vecfwd[c]{h}_{i,j}$ and $\vecrev[c]{h}_{i,j} \in \R^{256}$,
$\vecfwd[w]{h}_{i}$ and $\vecrev[w]{h}_{i} \in \R^{256}$,
$\mat[eh]{W} \in \R^{512 \times 256}$, $\vec{b}_{h} \in \R^{256}$,
$\mat[hy]{W} \in \R^{256 \times 1}$, $b_y \in \R$,
dropout rate to the embedding output: $0.5$, 
batch size: $32$. Training is performed until 100 epochs are reached.
The $RMSE_{user}$ results are shown in \tabref{tbl:performanceuserlevel} where EXT, STA, AGR, CON and OPN are abbreviations for Extroversion, Emotional Stability (the inverse of Neuroticism), Agreeableness, Conscientiousness and Openness respectively.

\myparagraph{\system{C2W2S4PT} outperforms the current state of the art in EN and ES} In the $5$-fold cross-validation group, \system{C2W2S4PT}  is superior to the baselines, achieving better performance except for CON in ES. In terms of the performance measured by $10$-fold cross-validation, the dominance of the proposed model is even more pronounced with \system{C2W2S4PT} outperforming the two selected baseline systems across all personality traits.
Overall, in comparison to the previous state-of-the-art models in both groups, \system{C2W2S4PT} not only outperforms them, by a significant margin in the case of $10$-fold cross-validation, but it also achieves so without any hand-crafted features, underlining the soundness of the approach.

\myparagraph{On CON in ES, $5$-fold cross-validation} We suspect that the surprisingly good performance of \newcite{Sulea+:2015} may likely be attributed to overfitting. Indeed, the performance on the test set on CON in ES is even inferior to \newcite{nowson2015}, further confirming our speculation. 

\myparagraph{The superiority of \system{C2W2S4PT} is less clear in IT} This can possibly be caused by the inadequate amount of Italian data, less than $4$k tweets as compared to $14$k and $10$k in the English and Spanish datasets, limiting the capability of \system{C2W2S4PT} to learn a reasonable model.

%Additionally, the model jointly trained on two personality traits, STA and AGR, the strongest correlated pair, achieves similar performance to its counterparts trained on each trait individually. Even the model jointly trained on a weakly correlated pair, STA and CON, generates competitive performance.

\begin{table*}[tb]
\center
\begin{tabular}{cc|l|ccccc}
%N-fold & \multirow{2}{*}{Model} & \multirow{2}{*}{EXT} & \multirow{2}{*}{STA} & \multirow{2}{*}{AGR} & \multirow{2}{*}{CON} & \multirow{2}{*}{OPN} \\
%CV & & & & & & \\
\hline
Lang. & $k$ & Model & EXT & STA & AGR & CON & OPN \\
\hline
\multirow{7}{*}{EN} & --- & \system{Average Baseline} & 0.166 & 0.223 & 0.158 & 0.151 & 0.146 \\
[1ex]
& \multirow{2}{*}{5} & \newcite{Sulea+:2015} & 0.136 & 0.183 & 0.141 & 0.131 & 0.119 \\
&  & \system{C2W2S4PT} & \textbf{0.131} & \textbf{0.171} & \textbf{0.140} & \textbf{0.124} & \textbf{0.109}\\
% & \system{C2W2S4PT - Multitask STA\&AGR} & \NA & 0.172 & 0.140 & \NA & \NA\\
% & \system{C2W2S4PT - Multitask AGR\&CON} & \NA & \NA & \textbf{0.138} & \textbf{0.124} & \NA\\
% & \system{C2W2S4PT - Multitask All} & 0.136 & 0.177 & 0.141 & 0.128 & 0.117\\
[1ex]
& \multirow{3}{*}{10} & \newcite{Mirkin+:2015} & 0.171 & 0.223 & 0.173 & 0.144 & 0.146\\
&  & \newcite{nowson2015} & 0.153 & 0.197 & 0.154 & 0.144 & 0.132\\
&  & \system{C2W2S4PT} & \textbf{0.130} & \textbf{0.167} & \textbf{0.137} & \textbf{0.122} & \textbf{0.109}\\
% & \system{C2W2S4PT - Multitask STA\&AGR} & \NA & 0.168 & 0.140 & \NA & \NA\\
% & \system{C2W2S4PT - Multitask AGR\&CON} & \NA & \NA & 0.138 & 0.123 & \NA\\
% & \system{C2W2S4PT - Multitask All} & 0.136 & 0.175 & 0.140 & 0.127 & 0.115\\
\hline
\multirow{7}{*}{ES} & --- & \system{Average Baseline} & 0.171 & 0.203 & 0.163 & 0.187 & 0.166\\
[1ex]& \multirow{2}{*}{5} & \newcite{Sulea+:2015} & 0.152 & 0.181 & 0.148 & \bf 0.114 & 0.142\\
&  & \system{C2W2S4PT} & \bf 0.148 & \bf 0.177 & \bf 0.143 & 0.157 & \bf 0.136\\
[1ex]
& \multirow{3}{*}{10} & \newcite{Mirkin+:2015} & 0.153 & 0.188 & 0.155 & 0.156 & 0.160\\
&  & \newcite{nowson2015} & 0.154 & 0.188 & 0.155 & 0.168 & 0.160 \\
&  & \system{C2W2S4PT} & \bf 0.145 & \bf 0.177 & \bf 0.142 & \bf 0.153 & \bf 0.137\\
\hline
\multirow{7}{*}{IT} & --- & \system{Average Baseline} & 0.162 & 0.172 & 0.162 & 0.123 & 0.151\\
[1ex]& \multirow{2}{*}{5} & \newcite{Sulea+:2015} & \bf 0.119 & 0.150 & 0.\bf 122 & 0.101 & \bf 0.130\\
&  & \system{C2W2S4PT} & 0.124 & \bf 0.144 & 0.130 & \bf 0.095 & 0.131\\
[1ex]
& \multirow{3}{*}{10} & \newcite{Mirkin+:2015} & \bf 0.095 & 0.168 & 0.142 & 0.098 & 0.137\\
&  & \newcite{nowson2015} & 0.137 & 0.168 & 0.142 & 0.098 & 0.141\\
&  & \system{C2W2S4PT} & 0.118 & \bf 0.147 & \bf 0.128 & \bf 0.095 & \bf 0.127\\
\hline
\end{tabular}
\caption{\label{tbl:performanceuserlevel}$RMSE_{user}$ across five traits. \textbf{Bold} highlights best performance.}
\end{table*}

\subsection{Personality Trait Prediction at Single Tweet Level}
\label{sec:tweetlevel}

%We test the proposed models on the dataset described in \secref{sec:data} and train our model to predict the personality trait score of each tweet based purely on the text with no additional features supplied. 
Although user-level evaluation is the common practice, we choose tweet-level performance to study the models' capabilities to infer personality at a lower granularity level. To support our evaluation, a number of baselines were created. To facilitate fair comparison, the only feature used is the surface form of the text. 
\system{Average Baseline} assigns the average of all the scores to each tweet. 
Also, two BoW systems, namely, \system{Random Forest} and \system{SVM Regression}, have been implemented for comparison. For these two BoW-based baseline systems, we perform grid search to find the best hyper-parameter configuration. For \system{SVM Regression}, the hyper-parameters include: kernel $\in \{\textrm{linear}, \textrm{rbf}\}$ and $C \in \{0.01, 0.1, 1.0, 10.0\}$ whereas for \system{Random Forest}, the number of trees is chosen from the set $\{10, 50, 100, 500, 1000\}$.

Additionally, two simpler RNN-based models, namely \system{Bi-GRU-Char} and \system{Bi-GRU-Word}, which only work on character and word level respectively but share the same structure of the final MLP classifier ($\vec{h}_{s}$ and $\hat{y}_s$), have also been presented in contrast to the more sophisticated character to word compositional model \system{C2W2S4PT}. For training, \system{C2W2S4PT} inherits the same hyper-parameter configuration as described in \secref{sec:userlevel}.
%All the RNN-based models are trained with Adam \cite{Kingma+:2014}. Other hyper-parameters include: 
%$\mat[c]{E} \in \R^{50 \times |C|}$, 
%$\vecfwd[c]{h}_{i,j}$ and $\vecrev[c]{h}_{i,j} \in \R^{256}$,
%$\vecfwd[w]{h}_{i}$ and $\vecrev[w]{h}_{i} \in \R^{256}$,
%$\mat[eh]{W} \in \R^{512 \times 256}$, $\vec{b}_{h} \in \R^{256}$,
%$\mat[hy]{W} \in \R^{256 \times 1}$, $b_y \in \R$,
%dropout rate to the embedding output: $0.5$, 
%batch size: $32$. Training is performed until 100 epoches are reached. 
For \system{Bi-GRU-Char} and \system{Bi-GRU-Word}, we set the character and word embedding size to $50$ and $256$ respectively. Due to time constraints, we did not perform hyper-parameter fine-tuning for the RNN-based models and \system{C2W2S4PT}. The $RMSE_{tweet}$ of each effort, measured by $10$-fold stratified cross-validation, is shown in \tabref{tbl:performancetweetlevel}.

\begin{table*}[tb]
	\center
%	\resizebox{\textwidth}{!}{
	\begin{tabular}{c|l|ccccc}
		\hline
		Lang. & Model & EXT & STA & AGR & CON & OPN \\
		\hline
		%\system{Linear Regression} & 0.562 & 0.670 & 0.542 & 0.529 & 0.400 \\
		\multirow{7}{*}{EN} & \system{Average Baseline} & 0.163 & 0.222 & 0.157 & 0.150 & 0.147 \\
		[1ex]
		& \system{SVM Regression} & 0.148 & 0.196 & 0.148 & 0.140 & 0.131  \\
		& \system{Random Forest} & 0.144 & 0.192 & \bf 0.146 & 0.138 & 0.132 \\
		%\system{Random Forest - Multitask STA\&AGR} & \NA & 0.199 & 0.150 & \NA & \NA \\
		%\system{Random Forest - Multitask AGR\&CON} & \NA & \NA & 0.150 & 0.145 & \NA \\
		%\system{Random Forest - Multitask All} & 0.151 & 0.201 & 0.150 & 0.145 & 0.135 \\
		[1ex]
		%\system{Char (LSTM)} & 0.152 & -- & -- & -- & --\\
		%\system{Word (LSTM)} & 0.145 & -- & -- & -- & --\\
		& \system{Bi-GRU-Char} & 0.150 & 0.202 & 0.152 & 0.143 & 0.137\\
		& \system{Bi-GRU-Word} & 0.147 & 0.200 & \textbf{0.146} & 0.138 & 0.130 \\
		%\system{Char+Word (LSTM)} & 0.143 & 0.193 & 0.149 & 0.138 & 0.129\\
		[1ex]
		& \system{C2W2S4PT} & \textbf{0.142} & \textbf{0.188} & 0.147 & \textbf{0.136} & \textbf{0.127}\\
		%\system{C2W2S4PT - Multitask STA\&AGR} & \NA & \textbf{0.189} & \textbf{0.146} & \NA & \NA\\
		%\system{Char+Word (Multitask STA\&CON)} & \NA & \textbf{0.188} & \NA & 0.137 & \NA \\
		%\system{C2W2S4PT - Multitask AGR\&CON} & \NA & \NA & \textbf{0.146} & \textbf{0.136} & \NA \\
		%\system{Char+Word (Multitask EXT\&STA)} & \textbf{0.142} & 0.190 & \NA & \NA & \NA \\
		%\system{C2W2S4PT - Multitask All} & \textbf{0.142} & 0.191 & \textbf{0.146} & 0.137 & 0.127 \\
		\hline
		
		\multirow{7}{*}{ES} & \system{Average Baseline} & 0.171 & 0.204 & 0.163 & 0.187 & 0.165\\
		[1ex]
		& \system{SVM Regression} & \bf 0.158 & \bf 0.190 & 0.157 & 0.171 & 0.152 \\
		& \system{Random Forest} & 0.159 & 0.195 & 0.157 & 0.177 & 0.158 \\
		[1ex]
		& \system{Bi-GRU-Char} & 0.163 & 0.195 & 0.158 & 0.178 & 0.155\\
		& \system{Bi-GRU-Word} & 0.159 & 0.192 & 0.154 & 0.173 & 0.154\\
		[1ex]
		& \system{C2W2S4PT} & \bf 0.158 & 0.191 & \bf 0.153 & \bf 0.168 & \bf 0.150\\
		\hline
		\multirow{7}{*}{IT} & \system{Average Baseline} & 0.164 & 0.171 & 0.164 & 0.125 & 0.153\\
		[1ex]
		& \system{SVM Regression} & 0.141 & 0.159 & 0.145 & 0.113 & \bf 0.141 \\
		& \system{Random Forest} & 0.140 & 0.161 & \bf 0.140 & 0.111 & 0.147\\
		[1ex]
		& \system{Bi-GRU-Char} & 0.149 & 0.163 & 0.153 & 0.117 & 0.146\\
		& \system{Bi-GRU-Word} & \bf 0.135 & \bf 0.156 & \bf 0.140 & \bf 0.109 & \bf 0.141\\
		[1ex]
		& \system{C2W2S4PT} & 0.139 & \bf 0.156 & 0.143 & \bf 0.109 & \bf 0.141\\
		\hline
	\end{tabular}
%	}
	\caption{\label{tbl:performancetweetlevel}$RMSE_{tweet}$ across five traits level. \textbf{Bold} highlights best performance.}
\end{table*}

%\myparagraph{RNN-based models outperform the three baselines} It can be observed that even the simplest RNN-based models, \system{Bi-GRU-Char} and \system{Bi-GRU-Word}, achieve comparable performance with the best baseline \system{Random Forest}. %, especially \system{Bi-GRU-Word} narrowly beating it by an average margin of $0.003$. 
%The strong performance of \system{Random Forest} can be ascribed to its ensemble nature, lowering the variance by taking multiple prediction results into account. 
%%The relatively incompetent performance of \system{Bi-GRU-Char} is likely caused by the rather lengthy character sequences in tweets \scott{cite a paper for this? comparison with shorter spans, eg pos-tagging maybe?}. RNNs, even LSTMs and GRUs, struggle to memorize information over long sequences \felix{cite papers}. 
%A further decrease in $RMSE_{tweet}$ is achieved by the more sophisticated \system{C2W2S4PT} model which is state of the art in almost every trait with the exception of AGR. This success can be attributed to the model's capability of coping with arbitrary words while not forgetting information due to excessive lengths as can arise from representing a text as a sequence of characters.
%%not having to deal with excessively long sequences. 
%It is impressive to achieve this with no feature engineering and that \system{C2W2S4PT}, a single learning algorithm, outperforms the ensemble method \system{Random Forest}.

\myparagraph{\system{C2W2S4PT} achieves comparable or better performance with \system{SVM Regression} and \system{Random Forest} in EN and ES} \system{C2W2S4PT} is state of the art in almost every trait with the exception of AGR in EN and STA in ES. This demonstrates that \system{C2W2S4PT} generates at least reasonably comparable performance with \system{SVM Regression} and \system{Random Forest} in the feature-engineering-free setting on the tweet level and it does so without exhaustive hyper-parameter fine-tuning.
%On the Spanish tasks, the gap in $RMSE_{tweet}$ between \system{C2W2S4PT} and \system{SVM Regression} and \system{Random Forest} is narrower. This can possibly be caused by the relatively smaller size of the Spanish dataset with less than 10k tweets whereas more than 14k tweets are included in the English set.

\myparagraph{\system{C2W2S4PT} outperforms the RNN-based baselines in EN and ES} This success can be attributed to the model's capability of coping with arbitrary words while not forgetting information due to excessive lengths as can arise from representing a text as a sequence of characters. Also, given that \system{C2W2S4PT} does not need to maintain a large vocabulary embedding matrix as in \system{Bi-GRU-Word}, there are much fewer parameters for the model to learn \cite{Ling+:2015}, making it less prone to overfitting.

\myparagraph{The performance of \system{C2W2S4PT} is inferior to \system{Bi-GRU-Word} in IT} 
%This can possibly be caused by the inadequate amount of Italian data, less than $4$k tweets as compared to $14$k and $10$k in the English and Spanish datasets, limiting the capability of \system{C2W2S4PT} to learn a reasonable model.
\system{Bi-GRU-Word} achieves the best performance across all personality traits with \system{C2W2S4PT} coming in as a close second and tying in 3 traits.
Apart from the inadequate amount of Italian data causing the fluctuation in performance as explained in \secref{sec:userlevel}, further investigation is needed to analyse the strong performance of \system{Bi-GRU-Word}.

%\myparagraph{Multitask learning provides little benefits to performance} Surprisingly, the model jointly trained on the weakest correlated pair, namely AGR\&CON, achieves even better results than the one trained on the strongest correlated pair (STA\&AGR). 
%%This could possibly be attributed to that the weakly correlated pair works as a mutual regularization term to prevent the model from overfitting to one particular personality trait, thereby achieving superior generalization performance. 
%In fact, despite the noise introduced by the training on non-correlated personality traits, there is little impact on the performance of the multitask-learning models and the model jointly trained on all 5 personality traits generates equally competitive performance.

%\begin{figure}[tb]
%%\vspace{-0.5cm}
%\begin{center}
%\resizebox{.9\textwidth}{!}{
%\input{resmemn2n.tex}
%}
%\end{center}
%%\vspace{-1.0cm}
%\caption{\label{fig:resmemn2n}Illustration of the proposed model with $3$ hops.}
%\end{figure}

\subsection{Visualisation}

%In order to investigate into the features automatically learned by the models, we propose to visualize the sentences using t-SNE \cite{Van+:2008} or PCA \cite{Tipping+:1999}. Specifically, we adopt a trained \system{Char+Word (GRU)} model to first construct word representations using the character to word compositional model and then construct sentence representations by feeding the word representations into a word-level RNN. Ultimately, sentences are represented by high dimensional scalar vectors. Next, we apply t-SNE or PCA to reduce the dimensionality of the representations so that they can be plotted into a scatter plot to better examine the patterns of features learned by the models. In \figref{fig:pca}, we show a scatter plot of sentence representations processed by PCA. 

To further investigate into the learned representations and features, we choose the \system{C2W2S4PT} model trained on a single personality trait and visualise the sentences with the help of PCA \cite{Tipping+:1999}. We also experimented with t-SNE \cite{Van+:2008} but it did not produce an interpretable plot. 100 tweets have been randomly selected (50 tweets each from either end of the EXT spectrum) with their representations constructed by the model. \figref{fig:pca} shows the scatter plot of the representations of the sentences reduced to a $2$D space by PCA for the trait of Extraversion (EXT), selected as it is the most commonly studied and well understood trait.
The figure shows clusters of both positive and negative Extraversion, though the former intersect the latter. For discussion we consider three examples as highlighted in \figref{fig:pca}:
%\vspace{-8mm}
\begin{figure}[tb]
	\center
	\includegraphics[width=0.5\textwidth]{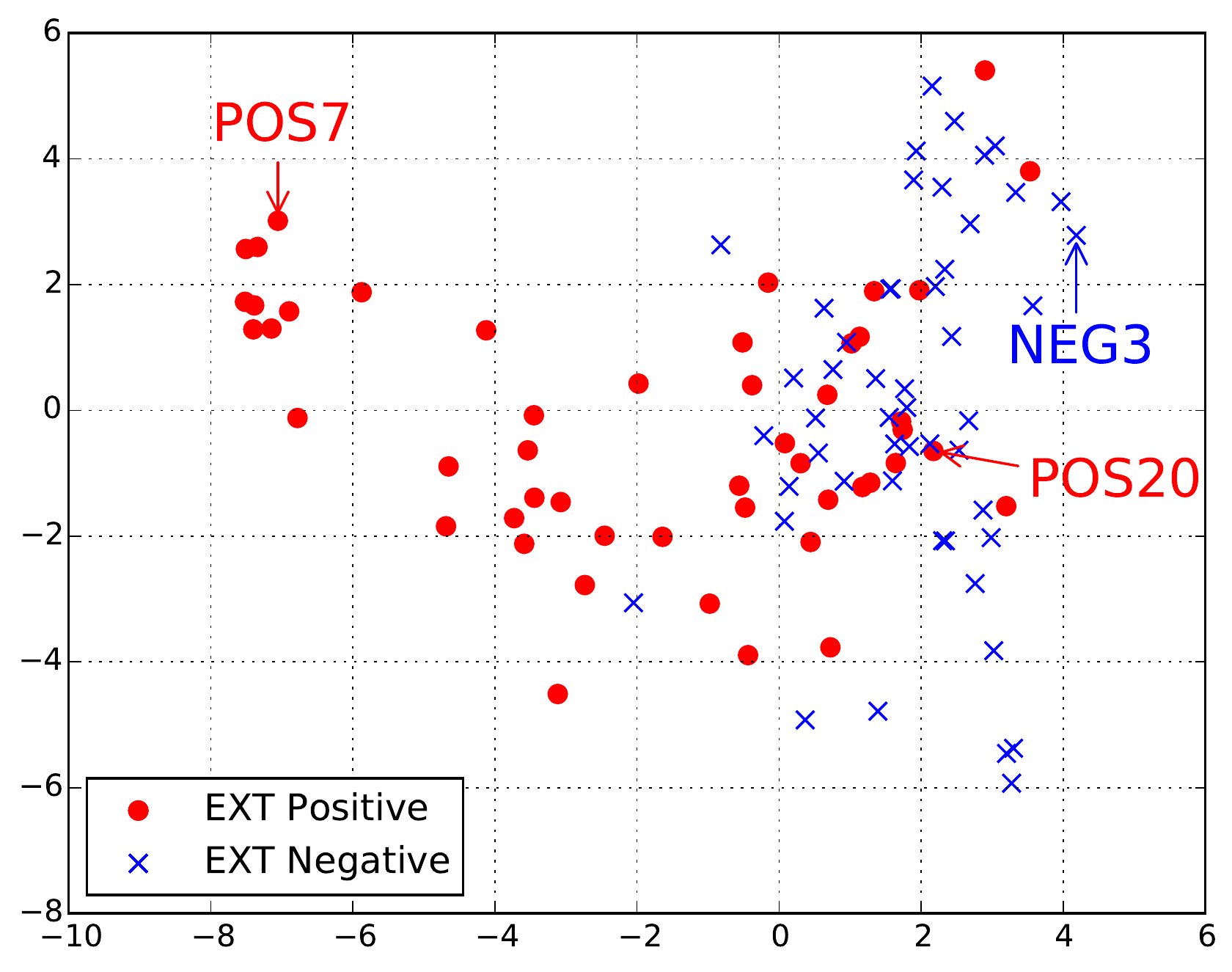}
%	\vspace{-4mm}
	\caption{\label{fig:pca}Scatter plot of sentence representations processed by PCA. }
%	\vspace{-6mm}
\end{figure}

\begin{itemize}[noitemsep,topsep=0pt]
	\itemsep0em
	\item POS7: \textit{``@username: Feeling like you're not good enough is probably the worst thing to feel.''}
	\item NEG3: \textit{``Being good ain't enough lately.''}
	\item POS20: \textit{``o.O Lovely.''}
\end{itemize}
%\vspace{2mm}
The first two examples (POS7 and NEG3) are drawn from largely distinct areas of the distribution. In essence the semantics of the short texts are the same. However, they both show linguistic attributes commonly understood to relate to Extraversion~\cite{gill02}: POS7 is longer and, with the use of the second person pronoun, is more inclusive of others; NEG3 on the other hand is shorter and self-focused, aspects indicative of Introversion.
The third sentence, POS20, is a statement from an Extravert which appears to map to an Introvert space. Indeed, while short, the use of ``Eastern'' style, non-rotated emoticons (such as \textit{o.O}) has also been shown to relate to Introversion on social media~\cite{schwartz2013}.
This is perhaps not the venue to consider the implications of this further, although one explanation might be that the model has uncovered a flexibility often associated with Ambiverts~\cite{grant2013}. However, it is important to consider that the model is indeed capturing well-understood dimensions of language yet with no feature engineering.

%\begin{figure}[h]
%\center
%\includegraphics[width=0.6\textwidth]{sent_pca.pdf}
%\vspace{-1cm}
%\caption{\label{fig:pca}Scatter plot of sentence representations processed by PCA. }
%\end{figure}

\section{Conclusion and Future Work}

%Overall, the results in the paper support our methodology: the \system{C2W2S4PT} model not only performs well on the short text level as we had intended, but also provides state-of-the-art results when adapted to user-level. However, interpretation of the performance of the multitask experiments is less straightforward. At text level (as per \tabref{tbl:performancetweetlevel}) the results are almost identical whether modelling traits individually, all together, or with differing prior relationships. Perhaps it is the case that simple linear correlations do not adequately explain the relationships between traits when mediated via language use. It could also be that our model captures a more complex, non-linear relationship or some notion of latent variables.  It is clear that this requires further investigation, though this will likely require an additional dataset, as with only 150 authors, the distribution of scores is somewhat limited.
%One advantage of our approach which requires validation is that lack of feature engineering should support language independence. Preliminary tests on the Spanish data from the PAN 2015 Author Profiling dataset show promising results. However, due to time constraints, we leave this exercise for future work.
%%Also, we plan to incorporate attention mechanisms, enabling us to visualize the focus of the model and thereby allowing for further investigation into the connections between one's personality and the choice of words.

Overall, the results in the paper support our methodology: \system{C2W2S4PT} not only provides state-of-the-art results on the user level, but also performs reasonably well when adapted to the short text level compared to other widely used models in the feature-engineering-free setting. More importantly, one advantage of our approach is the lack of feature engineering which allows us to adapt the same model to other languages with no modification to the model itself. To further examine this property of the proposed model, we plan to adopt TwiSty \cite{Verhoeven+:2016}, a recently introduced corpus consisting of 6 languages and labelled with MBTI type indicators \cite{Myers+:2010}.

\Urlmuskip=0mu plus 1mu\relax
\bibliographystyle{eacl2017}
\bibliography{strings-shrt,pmt-persproj,refs_personality,ml}

\end{document}